%% file: ijcai26.tex
\documentclass{article}
\usepackage{ijcai26}

\usepackage{times}
\usepackage{soul}
\usepackage{url}
\usepackage[hidelinks]{hyperref}
\usepackage[utf8]{inputenc}
\usepackage[small]{caption}
\usepackage{graphicx}
\usepackage{amsmath}
\usepackage{amsthm}
\usepackage{booktabs}
\usepackage{algorithm}
\usepackage{algorithmic}
\usepackage[switch]{lineno}

\nolinenumbers

\title{On the Generalization Gap in LLM Planning: Tests and Verifier-Reward RL}

\author{
Valerio Belcamino$^1$
\and
Nicholas Attolino$^1$\and
Alessio Capitanelli$^2$\And
Fulvio Mastrogiovanni$^1$\\
\affiliations
$^1$Department of Informatics, Bioengineering, Robotics and Systems Engineering, \\University of Genoa, Viale Causa 13, 16145 Genoa, Italy; \\
$^2$AIKO S.r.l., Via dei Mille 22, 10123, Torino, Italy\\
\emails
alessio.capitanelli@dibris.unige.it
}

\begin{document}

\maketitle

\begin{abstract}
Recent work shows that fine-tuned Large Language Models (LLMs) can achieve high valid plan rates on PDDL planning tasks.
However, it remains unclear whether this reflects transferable planning competence or domain-specific memorization. 
In this work, we fine-tune a 1.7B-parameter LLM on 40,000 domain–problem–plan tuples from 10 IPC 2023 domains, and evaluate both in-domain and cross-domain generalization. 
While the model reaches 82.9\% valid plan rate in \textit{in-domain} conditions, it achieves 0\% on two \textit{unseen} domains. 
To analyze this failure, we introduce three diagnostic interventions, namely
(i) instance-wise symbol anonymization, 
(ii) compact plan serialization, and 
(iii) verifier-reward fine-tuning using the VAL validator as a success-focused reinforcement signal. 
Symbol anonymization and compact serialization cause significant performance drops despite preserving plan semantics, thus revealing strong sensitivity to surface representations. 
Verifier-reward fine-tuning reaches performance saturation in half the supervised training epochs, but does not improve cross-domain generalization.
For the explored configurations, in-domain performance plateaus around 80\%, while cross-domain performance collapses, suggesting that our fine-tuned model relies heavily on domain-specific patterns rather than transferable planning competence in this setting.
Our results highlight a persistent generalization gap in LLM-based planning and provide diagnostic tools for studying its causes.
\end{abstract}

\section{Introduction}
\label{sec:intro}

Large Language Models (LLMs) have recently shown promising results on symbolic planning tasks formulated using the well-known Planning Domain Definition Language (PDDL) and its variants, especially after supervised fine-tuning on domain-specific datasets. 
Their ability to process structured input and to generate syntactically valid plans seems to suggest a potential role as \textit{learned} planners. 
However, strong \textit{in-domain} performance does not necessarily imply that LLMs can acquire transferable planning competence. 
On the contrary, they may rely on memorizing lexical patterns, regularities, or domain-specific action templates that could fail under distribution shifts.

We argue that understanding whether LLMs learn abstract planning principles, or merely domain-bound heuristics, is critical for their deployment in open-ended planning scenarios. 
Prior work has largely focused on improving in-domain performance through better prompting, smart synthetic data generation, or fine-tuning pipelines.
However, systematic analyses of cross-domain generalization remain limited in the literature.
In this work, we address the following research question:
\textit{To what extent do fine-tuned LLMs acquire transferable planning competence across PDDL-compatible domains, or do they primarily learn domain-specific superficial regularities?}
To address this problem, we 
(i) fine-tune a 1.7B-parameter LLM on 40,000 domain–problem–plan tuples spanning 10 domains from the International Planning Competition (IPC) \cite{taitler-et-al-aimag2024}, and 
(ii) evaluate its performance both on held-out in-domain instances and on two entirely unseen domains. Despite achieving 82.9\% valid plan rate in-domain, the model achieves 0\% on unseen domains, indicating a severe lack of cross-domain generalization under our setting.
Then, we further analyze this result through three \textit{variants}.
In variant \textsc{v1} we perform an instance-wise symbol anonymization to remove meaningful object, predicate, action names while preserving structural relations among them, thereby we isolate the role of \textit{lexical} semantics.
In \textsc{v2} we adopt compact plan serialization to remove \textit{superficial} formatting elements without changing plan semantics to ascertain sensitivity to surface representation.
In \textsc{v3} we introduce \textit{verifier-reward fine-tuning}, in which the VAL plan validator provides success-focused reinforcement signals based on functional correctness rather than string matching.

Our analyses lead to three key findings.
First, symbol anonymization and compact serialization cause large performance drops, suggesting that current LLM-based planners are highly sensitive to superficial representations.
Second, verifier-reward fine-tuning reaches performance saturation in roughly half the supervised training epochs, suggesting that functional validation can serve as an alternative optimization signal once a supervised warm start is available.
Third, none of these interventions improve cross-domain generalization, as all variants completely fail on unseen domains.
These results suggest that in this configuration our fine-tuned models mainly learn domain-specific patterns rather than demonstrating \textit{abstract} planning competence across domains. We provide a multi-domain benchmark, diagnostic stress tests, and a verifier-reward framework to support future investigations.

\section{Related Work}
\label{sec:related}

\noindent
\textit{LLMs, Reasoning, and the Illusion of Competence}.
LLMs often show strong performance on reasoning benchmarks, especially when intermediate steps are elicited, yet a growing body of work argues that these gains do not necessarily reflect robust logical competence under increased complexity or distribution shifts \cite{illusion-of-thinking,karan2025reasoning,yue2025does}. To mitigate these issues, structured prompting and neurosymbolic approaches explicitly separate logical structure from surface form and rely on external tools for verification \cite{symbcot,ranaldi2025improving,10.1145/3748239.3748249,chen2025comparative}. These insights highlight a \textit{tension} between apparent reasoning and systematic generalization, and motivate evaluations that emphasize functional correctness rather than linguistic plausibility.

\noindent
\textit{LLMs for Automated Planning}.
LLM-based planning is attractive for its flexible natural-language interfaces but must satisfy strict consistency and executability constraints. Prompt-based approaches can generate plausible plans yet struggle with grounding errors, constraint tracking, and generalization as complexity grows \cite{song2023llm,valmeekam2022large,goebel2025can}. Hybrid pipelines use LLMs for high-level decomposition while delegating correctness to classical planners or validators \cite{liu2023llm+,wang2023plan}. Fine-tuning and instruction-tuning on PDDL domains improve in-domain executability \cite{ijcai2023p839,verma2025teaching}, and extensions to robotics emphasize scalability and streaming execution \cite{Capitanelli,attolino2025achieving}. However, recent systematic studies still report fragile behavior and limited transfer beyond trained domains \cite{zhu2025language,mendez2025systematic}.

\noindent
\textit{PDDL, Plan Validation, and Verification-Centric Pipelines}.
PDDL is the standard language for specifying planning domains and problems \cite{aeronautiques1998pddl}, with PDDL~2.1 extending expressivity to temporal constructs \cite{fox2003pddl2}. Plan validity is checked using VAL, which verifies that a candidate plan respects domain constraints and achieves the goal \cite{howey2004val}. In most LLM-based planning pipelines, VAL is used as an inference-time filter, repair mechanism, or dataset-cleaning tool \cite{liu2023llm+,Capitanelli,verma2025teaching,valmeekam2022large,goebel2025can}, and is rarely treated as a direct training signal. In contrast, we explicitly integrate VAL into the optimization loop to encourage functional plan correctness rather than mere string matching.

\noindent
\textit{Reinforcement Learning from Verifiable Rewards}.
Reinforcement learning from verifiable rewards (RLVR) leverages tasks where outcomes can be checked automatically, and has been applied to self-play curricula, multi-turn agents, tool-augmented reasoning, and long-context optimization \cite{zhao2025absolute,wang2025ragen,singh2025agentic,wang2025loongrl}. Such approaches can improve sampling efficiency, although gains do not always translate into stronger underlying reasoning ability \cite{karan2025reasoning,yue2025does}. For planning tasks, goal-conditioned RL has been used to refine LLM behavior without explicit search \cite{hong2025planning}, and targeted fine-tuning can outperform prompting when strict correctness constraints apply \cite{finetunee}. Building on these insights, we use VAL’s deterministic validation outcome as a \textit{success-focused} reinforcement signal for planning: unlike prior work that uses validation mainly for filtering or evaluation, we incorporate it directly into training, and study how this affects in-domain efficiency and cross-domain generalization.


\section{Dataset Generation}
\label{sec:dataset}

\begin{figure}[t]
\centering
\includegraphics[width=\columnwidth]{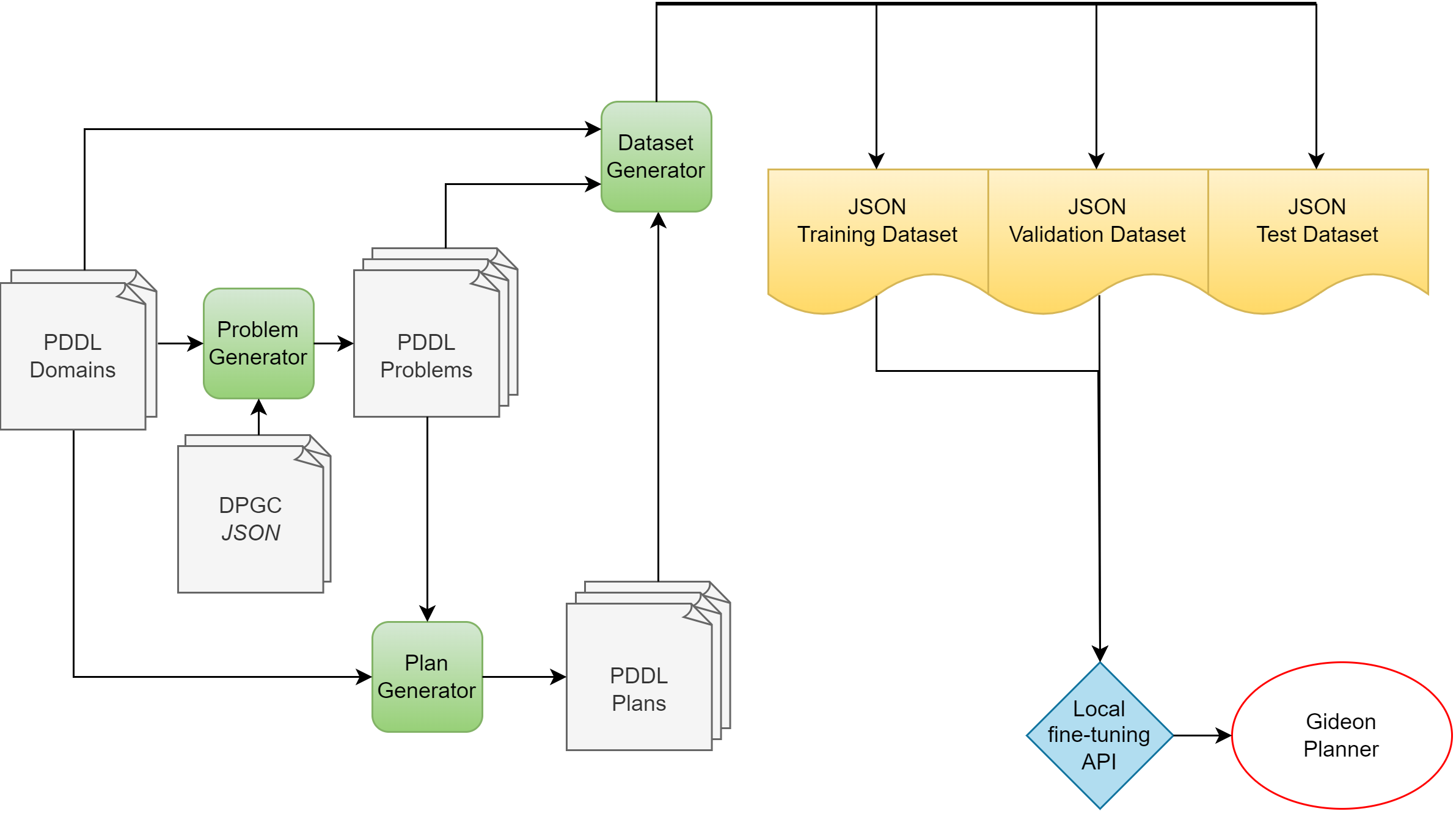}
\caption{Overview of the Gideon-based data generation pipeline and domain distribution.}
\label{fig:pipeline}
\end{figure}

In order to carry out our analysis, we leveraged the Gideon framework \cite{attolino2025achieving}, which is designed to systematically generate large-scale, high-quality datasets for neurosymbolic task planning.
Figure \ref{fig:pipeline} illustrates the process pipeline orchestrating the transformation of abstract PDDL domain specifications into validated domain-problem-plan tuples, which can be directly used to fine-tune LLMs for symbolic planning tasks. 
This pipeline enables the controlled generation of diverse, validated planning instances, supporting systematic analysis of generalization beyond small or hand-crafted benchmarks.

\noindent
\textit{Problem Generation}. 
The pipeline begins with a \textit{Problem Generator}, that takes as input one or more PDDL domains with the associated Domain–Problem Generation Configuration (DPGC) files. 
Each DPGC encodes the rules governing random problem generation, allowing users to explicitly control how initial and goal states are constructed while ensuring solvability and adherence to domain-specific conventions. 

Each generated problem separates domain invariants of the initial and final states from variable components, which are instantiated probabilistically. These variable components are defined through collections of predicate pools, in which predicates are sampled according to user-defined rules. Predicate arguments are drawn from object pools that specify object types, quantities, naming schemes, and selection modes, such as mutually exclusive or sequential usage. Synchronization mechanisms based on tagged references allow semantically related predicates to share or coordinate object selections. This is particularly useful in domains in which structural relations, for example related to topological or ordering constraints, must be preserved. 
Predicate pools can be mutually exclusive, which enables alternative but incompatible configurations to be selected according to predefined probabilities. 

Through these mechanisms, the Problem Generator module produces large numbers of diverse yet semantically coherent planning instances, avoiding trivial, unsolvable, or degenerate cases, and ensuring alignment with the domain's intended operational assumptions. 
The current implementation supports the core constructs of PDDL~2.1; general numeric fluents are excluded at this stage.

\noindent
\textit{Plan Generation}.
Once problem instances are generated, the pipeline invokes a \textit{Plan Generator} module, which is planner-agnostic and can interface with different classical planners. In our experiments, we use the satisficing planner Probe to compute a solution plan for each domain–problem pair. Every candidate plan is validated with VAL to ensure strict compliance with the PDDL semantics and goal achievement. When the planner’s native output format differs from VAL’s expectations, a conversion step is applied before validation. 
This validation stage is essential not only for dataset correctness, but also to guarantee that the plans used during training conform exactly to the syntactic and semantic constraints that the learned model is expected to reproduce at inference time.

\noindent
\textit{Dataset Generation}.
After plan generation and validation, the \textit{Dataset Generator} module assembles the resulting domain–problem–plan tuples into a unified dataset. The tuples are shuffled and partitioned into training, validation, and test subsets according to user-defined proportions. During this process, the pipeline enforces strict uniqueness: duplicate tuples are discarded to prevent dataset imbalance and to eliminate the risk of data leakage across splits. The final dataset thus contains only validated, non-duplicated examples that are directly usable for fine-tuning and downstream evaluation.

\noindent
\textit{Anonymization}.
To analyze the possible role of implicit semantic cues in symbol names (which we explore in variant \textsc{v1}), we extended Gideon with an Anonymization module aimed at systematically renaming all symbolic identifiers in domain, problem, and plan files. 
As we pointed out in the Introduction, the goal is to prevent the learned model from exploiting natural-language meanings of actions, predicates, or objects, and to focus the learning process on structural regularities instead.
At the same time, our instance-wise anonymization also breaks symbol identity across problems within the same domain, turning each tuple into a self-contained symbolic system. Therefore, this module allows for a controlled evaluation of \textit{how much} planning performance depends on lexical semantics and cross-instance symbol reuse.
For each domain-problem-plan tuple, all action names, predicate names, object identifiers, and types are replaced with synthetic symbols, for example \texttt{a\_0}, \texttt{p\_3}, \texttt{o\_7}. 
The renaming is applied consistently within a tuple but independently across tuples, ensuring that no semantic or cross-instance information can be recovered from symbol names alone. 
The transformation preserves the structure of the problem, so that all relations and constraints remain unchanged, and the resulting files remain valid PDDL specifications.

\noindent
\textit{Dataset Creation}.
The training and validation dataset covers 10 IPC 2023 domains, summarized in Table~\ref{tab:domains_metrics}.
Two additional domains, that is, \textit{Rover} and \textit{Briefcase}, are used exclusively to evaluate cross-domain generalization. 
All domains are publicly available\footnote{Link: Code and data will be made available on publication.}.

Because of limitations in the PDDL features supported by Gideon, we applied minor adaptations to some domains to ensure compatibility with the DPGC framework. 
For example, in the \textit{Child-snack} domain, we added \texttt{:action-costs} with related \texttt{:functions} according to the DPGC constraints. 
These adaptations preserve the original planning semantics and enable systematic problem generation.

Overall, we generated 40,000 tuples (that is, 4,000 per domain) for training and validation, and 500 tuples (that is, 250 per unseen domain) for cross-domain testing. 
Table~\ref{tab:domains_metrics} reports token-length statistics for each training domain, where only a small fraction of instances exceed the 4,096-token limit.

\begin{table}[t]
\centering
\footnotesize
\renewcommand{\arraystretch}{1.2}
\begin{tabular}{lcc}
\hline
\textbf{Domain} & \textbf{Tuple} $\mathbf{> 4096}$ & \textbf{Longest Tuple} \\
\hline
\hline
\textit{Ferry} & $0$ & $2085$ \\
\textit{Floor-tile}       & $0$  & $3772$\\
\textit{Blocksworld}       & $0$     & $2085$ \\
\textit{Child-snack}       & $0$ & $2573$ \\
\textit{Spanner}       & $0$ & $1198$ \\
\textit{Satellite}       & $0$   & $2708$ \\
\textit{Maintenance} & $0$ & $1058$ \\
\textit{Parking}           & $12$  & $5319$\\
\textit{Transport}           & $6$     & $4214$ \\
\textit{Miconic}           & $0$ & $2907$ \\
\hline
\end{tabular}
\caption{For each domain, the number of tuples exceeding the maximum token limit allowed in our setup, and the longest tuple.}
\label{tab:domains_metrics}
\end{table}

\section{Models and Training Setups}
\label{sec:models}

\begin{figure}[t]
\centering
\includegraphics[width=0.9\columnwidth]{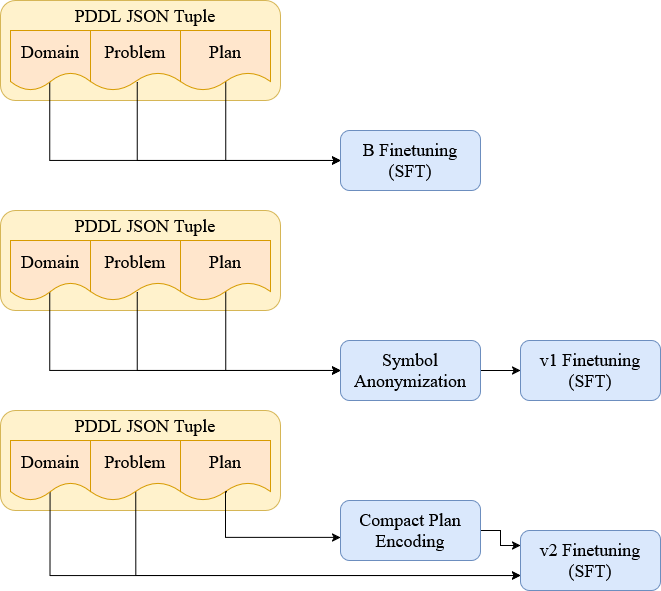}
\caption{
The three SFT pipelines. 
In all three cases, the model receives domain and problem as input, and the generated plan is compared to the reference plan produced by Probe. 
The diagram on the top shows how we get the baseline \textsc{B}.
The diagram in the mid illustrates the anonymization step for variant \textsc{v1}.
The diagram on the bottom shows the compact encoding stage that simplifies the syntax of the reference PDDL plans for variant \textsc{v2}.
}
\label{fig:sft_pipeline}
\end{figure}

All experiments use Qwen-3-1.7B, initialized from the official pretrained checkpoint on Hugging Face. Supervised fine-tuning (SFT) is performed with LLaMA-Factory. The baseline model, denoted \textsc{B}, and the SFT variants, denoted \textsc{v1} (instance-wise symbol anonymization) and \textsc{v2} (compact plan encoding) are each trained for three epochs, saving a checkpoint after every epoch. The three SFT training pipelines are illustrated in Figure \ref{fig:sft_pipeline}.

\begin{figure}[t]
\centering
\includegraphics[width=0.9\columnwidth]{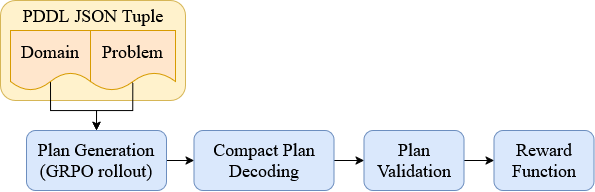}
\caption{
The RL pipeline used in this work.
The model takes a domain–problem pair and, via GRPO rollouts, generates multiple candidate plans, which are decoded (timestamps and brackets restored), validated by VAL, and scored.}
\label{fig:rl_pipeline}
\end{figure}

Reinforcement learning (RL) is applied on top of the 1-epoch \textsc{v2} checkpoint, which serves as the initialization for variant \textsc{v3} (verifier-reward fine-tuning). Directly initializing RL from the pretrained base model leads to extremely sparse rewards, whereas the 1-epoch checkpoint already produces sufficiently well-formed plans to provide a stable warm start. Alternative bootstrapping strategies are left to future work. Variant \textsc{v3} training pipeline is illustrated in Figure~\ref{fig:rl_pipeline}. 

Unless otherwise stated, all variants share the same optimizer configuration and maximum sequence length, ensuring a fair comparison across training setups.



\subsection{\textsc{v1}: Instance-wise Symbol Anonymization}
\label{sec:variant1}

Variant \textsc{v1} ascertains the model's reliance on \textit{semantically meaningful} symbol names. 
As we discussed in Section \ref{sec:dataset}, all action names, predicate names, and object identifiers are replaced with arbitrary symbols, while preserving arity, typing, and the relational structure.

To mitigate the abrupt loss of lexical cues, we adopt a curriculum learning strategy within a single training epoch.
We construct an expanded dataset by stacking the original dataset three times, resulting in training tuples indexed by $i = 1, \dots, N$. 
Each tuple is anonymized with a probability

\[
p(i) = \frac{i - 1}{N - 1},
\]
which increases linearly from 0 for the first tuple to 1 for the last one. 
As a result, early training samples tend to preserve original symbol names, while later examples are fully anonymized, therefore yielding a smooth transition from lexical grounding to purely structural supervision.
Then, training is performed for a single epoch over the expanded dataset. 

\subsection{\textsc{v2}: Compact Plan Encoding}
\label{sec:variant2}

Variant \textsc{v2} evaluates sensitivity to superficial plan formatting. 
Plan timestamps, parentheses, and the plan termination string \texttt{END} are removed, whereas the underlying sequence of actions is preserved. 
This compact encoding reduces token count during training without altering plan semantics.

For example, a trivial plan instance of the form:
\begin{quote}
\small
\begin{verbatim}
00100: (move truck1 depot1 depot2)
END
\end{verbatim}
\end{quote}
takes the following compact form:
\begin{quote}
\small
\begin{verbatim}
move truck1 depot1 depot2
\end{verbatim}
\end{quote}

Since VAL requires timestamps and parentheses, generated outputs are deterministically decoded into the standard format before validation occurs. 
This transformation is purely syntactic and does not modify the generated action sequence.
The encoded variant is trained for three epochs.

\subsection{\textsc{v3}: Verifier-Reward Fine-Tuning}
\label{sec:variant3}
As anticipated, variant \textsc{v3} training is initialized from the 1-epoch supervised checkpoint of \textsc{v2}, such that reinforcement operates on the compact (encoded) plan representation.
At each training step, the model receives as input the serialized PDDL domain and problem descriptions, and generates multiple candidate plans for the same instance. This multiplicity arises from the use of the Group Relative Policy Optimization (GRPO) algorithm, which stochastically samples a group of rollouts per problem to estimate relative advantages among candidate solutions.

Since the model operates on the encoded plan format, each generated plan is first deterministically decoded into the standard PDDL-compatible syntax expected by VAL. This decoding step restores timestamps, parentheses, and the plan termination marker, without altering the underlying action sequence.
Each decoded plan is then validated using VAL in \textit{verbose} mode. In this configuration, VAL not only determines whether a plan is valid and achieves the goal, but also reports the specific failure mode, such as unsatisfied preconditions or incomplete goal achievement. This fine-grained feedback is required to assign differentiated rewards.

Based on the validator output, a scalar reward is assigned to each candidate plan as:
$+1.0$ if the plan is valid \textit{and} the goal is reached;
$+0.1$ if the plan is syntactically valid but does not reach the goal;
$-0.1$ if execution fails due to unsatisfied preconditions;
$0.0$ for syntax errors or other failures.
GRPO uses these rewards to compute relative advantages within each rollout group, and update the model parameters accordingly.
RL is conducted for one additional epoch on top of the 1-epoch supervised checkpoint, after which performance saturates.

\section{Experimental Protocol}
\label{sec:protocol}

\noindent
\textit{Evaluation Metric}.
In this work, we adopt the \textit{valid plan rate} as the primary evaluation metric.
It is defined as the percentage of generated plans \textit{passing} VAL validation. 
A plan is considered \textit{valid} if and only if all action preconditions are satisfied, \textit{and} the goal state is achieved under the PDDL semantics. 
Plans that are syntactically correct but fail any VAL check receive no partial credit.

\noindent
\textit{Data Splits}.
As anticipated, we generate a total of 41,500 tuples across 10 IPC 2023 domains, with the following splits:
\begin{itemize}
\item \textit{Training/Validation}: 40,000 tuples (split ratio: 80/20).
\item \textit{In-domain test}: 1,000 tuples (100 per domain), held out from training.
\item \textit{Unseen-domain test}: 500 tuples (250 per domain) from two IPC domains not considered for training.
\end{itemize}
All sets are balanced across domains. Given the test set sizes, we report single-run results, which we consider sufficient for this exploratory study.

\begin{table}[t]
\centering
\footnotesize
\renewcommand{\arraystretch}{1.2}
\begin{tabular}{lll}
\hline
\textbf{Training ID} & \textbf{Parameter} & \textbf{Value} \\
\hline
\hline
\textsc{B} (SFT)    & Initialization        & Qwen-3-1.7B \\
\textsc{v1} (SFT)   & Learning rate         & $10^{-5}$ (cosine)\\
\textsc{v2} (SFT)   & Batch size            & $1$ \\
                    & Maximum input length  & $4096$ \\
                    & Gradient accumulation & $16$ \\
                    & Temperature           & $0.01$ \\
\hline
\hline
\textsc{v3} (GRPO)  & Initialization        & 1-epoch \textsc{v2} \\
                    & Learning rate         & $5 \times 10^{-6}$ (cosine)\\
                    & Batch size            & $1$ \\
                    & Maximum input length  & $4096$ \\
                    & Gradient accumulation & $8$ \\
                    & Temperature           & $0.5$ \\
                    & Top-p                 & $0.95$ \\
                    & Top-k                 & $1.0$ \\
                    & Rollout               & $3$ \\
\hline
\end{tabular}
\caption{
Summary of the training configurations adopted for the three variants considered in this work. 
The first column reports the variant name and training algorithm, while the remaining columns list the corresponding hyperparameters.}
\label{tab:training_config}
\end{table}

\noindent
\textit{Decoding Configuration}.
At inference time, a greedy decoding strategy is adopted for all experiments, with $\text{temperature}=0.01$, $\text{top-p}=1.0$, and no sampling. 
The maximum output length is set to 2048 tokens to minimize truncation of valid plans while respecting GPU memory constraints. 
Generations exceeding the length limit without completion are counted as failures.

\noindent
\textit{Training Configuration}.
Table~\ref{tab:training_config} summarizes the training configurations. 
All variants are initialized from the same Qwen-3-1.7B checkpoint. 
Supervised variants use next-token prediction with a batch size of $1$ and a learning rate of $10^{-5}$. 
Variant \textsc{v3} is initialized from the 1-epoch \textsc{v2} checkpoint and continues training with GRPO using a batch size of $1$, gradient accumulation of $8$, $\text{temperature}=0.5$, $\text{top-}p=0.95$, $\text{top-}k=1$, and a learning rate of $5 \times 10^{-6}$. 
For both SFT and GRPO the maximum input length is set to 4096. The analysis in Table \ref{tab:domains_metrics} illustrates that 8 out of 10 domains never exceed this limit.
The \textit{Parking} and \textit{Transport} domains exceed it in only 1.2\% and 0.6\% of the cases respectively.

\noindent
\textit{Fairness Controls}.
To ensure a valid comparison across variants: 
(i) \textsc{B}, \textsc{v1}, and \textsc{v2} are all initialized from the same pretrained checkpoint, while \textsc{v3} is bootstrapped from the 1-epoch \textsc{v2} checkpoint;
(ii) training, validation, and test splits are identical, with format transformations applied consistently; and 
(iii) all models are evaluated under identical decoding settings using VAL. For \textsc{v1}, anonymization is applied at both training and test time. 
In the case of \textsc{v2} and \textsc{v3}, evaluation is performed after decoding plans into the standard PDDL format.

\noindent
\textit{Compute Environment}.
Experiments run on a Kubeflow cluster with two pods, each hosting an NVIDIA RTX A6000 GPU (48~GB VRAM). Supervised fine-tuning uses a single pod, while GRPO training uses both pods with Fully Sharded Data Parallel (FSDP).

On this setup, \textsc{B} and \textsc{v1} (SFT) require approximately 54 hours for three epochs; \textsc{v2} (SFT) requires roughly 30\% less due to shorter input–output sequences. In contrast, \textsc{v3} takes roughly five times the wall-clock time to complete a single RL epoch due to repeated plan sampling, decoding, and validation, even though RL operates on the encoded plans used in \textsc{v2}. Thus, RL converges in fewer epochs but at a higher per-epoch cost. Both code and data are freely available\footnote{Link: Code and data will be made available on publication.}.


\section{Results}
\label{sec:results}

\subsection{Baseline In-Domain Performance}

Table~\ref{tab:indomain_results} reports per-domain valid plan rates on the 1,000-tuple in-domain test set. 
The supervised baseline \textsc{B} achieves 75.2\% after two epochs and 82.9\% after three. 
For comparison, \textsc{v1}, \textsc{v2}, and \textsc{v3} reach 71.4\%, 72.2\%, and 78.1\% average validity, respectively. Several domains reach near-perfect performance (\textit{Floor-tile}: 100\%, \textit{Spanner}: 100\%, \textit{Ferry}: 99\%), whereas \textit{Satellite} remains challenging, peaking at 16\%.

\begin{table}[t]
\centering
\footnotesize
\caption{
Per-domain valid plan rates (\%) on the in-domain test set (100 instances per domain) for the supervised baseline \textsc{B} (1--3 epochs) and variants \textsc{v1} (anonymization, 3 epochs), \textsc{v2} (compact encoding, 3 epochs), and \textsc{v3} (verifier-reward; 1 SFT + 1 RL epoch). 
}
\label{tab:indomain_results}
\begin{tabular}{lcccccc}
\hline
\textbf{Domain} & \textbf{\textsc{B}-1e} & \textbf{\textsc{B}-2e} & \textbf{\textsc{B}-3e} & \textbf{\textsc{v1}-3e} & \textbf{\textsc{v2}-3e} & \textbf{\textsc{v3}} \\
\hline
\hline
\textit{Ferry}           & 84 & 95 & 99 & 96 & 95 & 92 \\
\textit{Floor-tile}      & 99 & 95 & 100 & 93 & 73 & 80 \\
\textit{Blocksworld}     & 96 & 50 & 90 & 100 & 4 & 92 \\
\textit{Child-snack}     & 80 & 89 & 91 & 90 & 95 & 88 \\
\textit{Spanner}         & 58 & 100 & 100 & 98 & 100 & 99 \\
\textit{Satellite}       & 6 & 10 & 16 & 13 & 16 & 34 \\
\textit{Maintenance}     & 86 & 94 & 98 & 0 & 99 & 98 \\
\textit{Parking}         & 10 & 50 & 56 & 45 & 62 & 14 \\
\textit{Transport}       & 27 & 79 & 84 & 86 & 88 & 91 \\
\textit{Miconic}         & 74 & 90 & 95 & 93 & 90 & 93 \\
\hline
\hline
\textbf{Average} & 62.0 & 75.2 & 82.9 & 71.4 & 72.2 & 78.1 \\
\hline    
\end{tabular}
\end{table}

These results seem to confirm that SFT enables strong in-domain performance on several IPC domains, which is consistent with prior work on LLM-based planning. However, high average validity masks substantial domain heterogeneity, with some domains, such as \textit{Satellite} and \textit{Parking}, still difficult despite additional training epochs, and some domains (e.g., \textit{Blocksworld}) exhibiting non-monotonic behavior across epochs, suggesting interactions between token-level loss and plan validity.

\subsection{Unseen-Domain Generalization}

All variants are evaluated on 500 instances from two IPC domains, namely \textit{Rover} and \textit{Briefcase}, which are absent from the training set. 
It must be emphasized that, despite strong in-domain performance, both \textsc{B} and \textit{all} variants achieve 0\% valid plan rate on both unseen domains.

A qualitative inspection of VAL outputs and generated plans on these domains shows that the PDDL syntax is correct, actions are correctly parameterized, and violations of action preconditions are relatively rare. Instead, the model tends to get caught in loops or to wander without making progress toward the goal, for both the baseline and all variants (with examples available in the dataset repository). This suggests that the model has internalized some aspects of the domain dynamics (e.g., action parameters typing) but struggles to reliably steer behavior toward unseen goals in new domains. Given that we only evaluate two unseen domains with a single 1.7B model, we view this 0\% result as strong evidence of a severe generalization gap in our setting rather than a definitive statement about all fine-tuning paradigms.

\subsection{Diagnostic Variant Results}

\noindent
\textit{Instance-wise Symbol Anonymization} (\textsc{v1}).
This stress test achieves 71.4\% aggregate validity at three epochs, compared to 82.9\% for the baseline, corresponding to a drop of 11.5 percentage points. 
Most domains exhibit moderate degradation, but \textit{Maintenance} collapses from 98\% to 0\%, while others such as \textit{Blocksworld} slightly improve (100\% vs. 90\%).

\noindent
\textit{Compact Serialization} (\textsc{v2}).
Removing step identifiers while preserving the semantic content yields 72.2\% validity at three epochs. On average this is a drop of over 10 percentage points despite semantic equivalence, with strong domain variance: \textit{Transport} slightly improves (88\% vs. 84\% for \textsc{B}), whereas \textit{Blocksworld} collapses to 4\% (vs.\ 90\%).

\begin{figure}[t]
\centering
\includegraphics[width=\columnwidth]{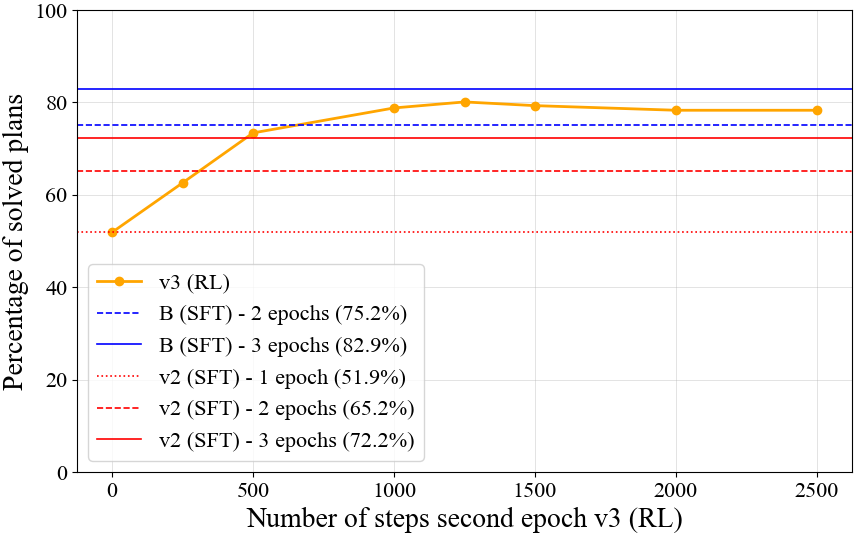}
\caption{
Variant \textsc{v3} reaches approximately 80\% validity in 1.5 total epochs (1 SFT + 0.5 RL).
This exceeds the 2-epoch supervised baseline and approaches the 3-epoch ceiling.}
\label{fig:rl_efficiency}
\end{figure}

\noindent
\textit{Verifier-Reward Training} (\textsc{v3}).
Figure~\ref{fig:rl_efficiency} compares training dynamics. 
Variant \textsc{v3}, initialized from a 1-epoch supervised checkpoint of \textsc{v2} (with a 51.9\% validity), reaches 80.1\% validity within 0.5 additional RL epochs. 
This exceeds the 2-epoch supervised baseline \textsc{B} (with a 75.2\% validity), and approaches the 3-epoch performance. 
Baseline SFT and verifier-reward RL plateau around 80--83\% in-domain validity, whereas anonymization and compact encoding plateau around 71--72\%. This suggests that the performance limit likely reflects factors beyond the training signal alone, such as model capacity, intrinsic task difficulty, or the chosen representations.
It is also noteworthy that \textsc{v3} nearly recovers the performance of the baseline \textsc{B} trained on the standard PDDL format, despite being initialized from a 1-epoch \textsc{v2} checkpoint and adopting the same compact plan encoding. This suggests that, while compact serialization is detrimental for pure supervised imitation (\textsc{v2} vs.\ \textsc{B}), a functional verifier-based reward can substantially mitigate this representational handicap for in-domain optimization, even though it does not improve cross-domain generalization.
Per-domain results in Table~\ref{tab:indomain_results} illustrate this heterogeneity: \textsc{v3} almost fully undoes the degradation of \textsc{v2} in \textit{Blocksworld} (92\% vs. 4\%) and substantially improves \textit{Satellite} (34\% vs.\ 16\%), but can hurt domains such as \textit{Parking} (14\% vs.\ 62\% for \textsc{v2}).

\subsection{Data Generation with Verifier-Reward}
\label{sec:cost}

Here, we analyze the computational implications of planner-based \textit{versus} verifier-reward supervision.
In particular, we focus on how the two strategies differ in dataset generation cost and training efficiency.

If we consider planner-based supervision, each training instance requires the generation of a reference plan using a classical planner.
This introduces a substantial upfront cost, which scales with problem difficulty and dominates dataset construction. 
In contrast, the training process leveraging the verifier-reward strategy eliminates the need for planner-generated labels, that is, they are replaced by repeated plan validation during training.

From a theoretical perspective, verifier-reward training shifts computational cost from dataset generation to training time. Although it requires sampling multiple candidate plans per problem and validating them, plan validation is polynomial in plan length and much cheaper than full planning, making the trade-off potentially favorable when scaling to larger datasets. Empirically, it requires substantially more computational resources and wall-clock time per epoch due to repeated sampling, decoding, and validation, yet converges in fewer epochs, reaching performance saturation faster than supervised training in terms of validation metrics per epoch. In our setup, however, the higher per-epoch cost means that the total wall-clock time to reach saturation with verifier-reward RL is not lower than that of a purely supervised run, even though it avoids generating planner labels for every training instance after the warm start. This explains why \textsc{v3} achieves comparable performance with fewer epochs despite higher per-step cost and overall higher wall-clock time per epoch.

A formal asymptotic analysis is provided in the Appendix (part of the Supplementary Material).
Therein, the total complexity of dataset generation, inference, verification, and training is derived for both paradigms. 
The Appendix further discusses how verifier-reward supervision may reduce reliance on planner-generated data and how the choice of the supervised bootstrapping checkpoint could be optimized.

\subsection{Sensitivity to Representation}

\noindent
\textit{Instance-wise Symbol Anonymization} (\textsc{v1}).
Replacing informative names with arbitrary symbols tests whether the model exploits lexical semantics, for example, inferring that \texttt{drive} implies \textit{movement}.
The observed 11.5\% drop confirms that lexical cues contribute to performance.
The domain-specific pattern is revealing: \textit{Maintenance} collapses from 98\% to 0\%, while \textit{Transport} and \textit{Blocksworld} remain stable or improve.
This heterogeneity suggests varying reliance on lexical cues and cross-instance symbol regularities across domains.

\noindent
\textit{Compact Serialization} (\textsc{v2}).
Removing step identifiers while preserving semantics yields a drop of more than 10\% on average.
This shows that formatting affects next-token prediction.
This sensitivity to superficial form, despite identical semantics, indicates that competence is tightly coupled to textual representations rather than the abstract plan \textit{structure}.

\noindent
\textit{Verifier-Reward Impact on Generalization} (\textsc{v3}).
Variant \textsc{v3} shows that binary feedback from VAL can partially substitute reference-plan supervision once a supervised warm start is available. 
Since the reward depends only on functional success rather than matching a specific action sequence, the model does not have any incentive to adhere to reference plans. 
This approach achieves comparable performance in roughly half the number of training epochs, albeit with higher per-epoch computational cost. Notably, although \textsc{v3} is initialized from a weaker 1-epoch \textsc{v2} checkpoint and continues to operate on the compact plan encoding, it nearly recovers the performance of the baseline \textsc{B} trained on the standard PDDL format, suggesting that a functional verifier-based reward can substantially mitigate the representational handicap of compact serialization for in-domain optimization.
However, the failure in improving cross-domain generalization 
suggests that the main bottlenecks for cross-domain transfer in our setting lie in factors such as model capacity, domain diversity, or representation choices rather than the choice of optimization signal alone.

\subsection{Limitations}

Several factors may help explain our results.

Both anonymization and compact serialization alter token-level inputs, and performance changes may partially reflect the tokenizer behavior rather than purely semantic effects. Furthermore, ten training domains and two unseen domains may be insufficient to induce abstract planning competence, as transferable planning behavior may emerge only with substantially greater domain diversity. Instance-wise anonymization impacts on two factors, namely the removal of semantic meaning, and the elimination of cross-instance symbol consistency. These effects cannot be uncorrelated in the current setup, although the catastrophic failure on \textit{Maintenance} may indicate that symbol identity plays a critical role at least in some domains. Verifier-reward training removes the explicit incentive to match reference strings. However, the model may still exploit surface regularities learned during the supervised warm-start phase. On the unseen domains, a qualitative inspection reveals that generated plans are mostly syntactically correct and respect action preconditions, yet often enter loops or wander without achieving the goal, suggesting that the main difficulty lies in reliably steering behavior toward unseen goals in new domains rather than in basic PDDL syntax or precondition satisfaction.

Finally, the observed plateau around 80--83\% validity for the strongest variants, and around 71--72\% under anonymization and compact encoding, may reflect model capacity limits, irreducible domain difficulty, training data ambiguity, or optimization dynamics.
A more thorough analysis of these factors will likely require larger models, broader domain coverage, or more structured datasets.
Overall, our results consistently indicate that high in-domain performance does not reliably translate into robust planning behavior on unseen domains in our setting, and that current LLM-based planners are strongly tied to surface-level representations and domain-specific regularities. 

\section{Conclusion}
\label{sec:conclusion}

This work investigates whether fine-tuned Large Language Models acquire transferable planning competence or primarily rely on superficial, token-level regularities. 

Starting from a benchmark of 40,000 domain-problem-plan tuples, generated using 10 domains borrowed from the IPC 2023, we show that a 1.7B-parameter model achieves high in-domain performance (82.9\% valid plan rate) but fails completely (0\%) on unseen domains.
We designed a few variants of the baseline model, to investigate the specific relevance of certain structural traits.
Symbol anonymization and compact plan serialization show that the model exhibits strong sensitivity to representational and lexical cues.
Semantically neutral transformations caused performance drops exceeding 10\%, which indicates that learned behavior is tightly coupled to superficial form rather than abstract planning structure.
We further showed how verifier-reward fine-tuning using VAL provides a viable alternative to reference-plan supervision. 
Such approach optimizes for functional correctness rather than string matching, and it reaches performance saturation in roughly half the number of training epochs.
However, while it improves label-efficiency and reduces dependence on planner-generated labels, it does not improve cross-domain generalization.

These results suggest that, in our setting, fine-tuning a 1.7B-parameter LLM on 10 IPC domains mainly induces domain-specific pattern learning rather than clearly demonstrating transferable planning competence.
Models appear to exploit lexical and structural regularities instead of abstracting the underlying plan semantics.
The label-efficiency gains enabled by verifier-reward training strategies make scaling to large and diverse datasets more tractable, even though per-epoch computational cost is higher in our setup.
The variants and the training framework introduced in this work could provide a foundation for future investigations into closing the gap between imitation \textit{at the surface} and functional generalization, with the aim of moving toward LLM-based planners that genuinely understand the problems they solve. In particular, we hope that the released multi-domain dataset, representation stress tests, and VAL-based RL pipeline will serve as practical tools for probing and improving generalization in future work under different settings and choices of models.

\bibliographystyle{named}
\bibliography{ijcai26}
\include{appendix_incl.tex}
\end{document}

%% file: appendix_incl.tex

\typeout{IJCAI--ECAI 26 Instructions for Authors}

\pdfpagewidth=8.5in
\pdfpageheight=11in

\linenumbers

\urlstyle{same}

\pdfinfo{
/TemplateVersion (IJCAI.2026.0)
}

\title{IJCAI--ECAI 26 Formatting Instructions}

\author{
    Author Name
    \affiliations
    Affiliation
    \emails
    email@example.com
}

\onecolumn
\appendix
\section{Computational Cost Analysis}
\label{app:cost}

In this Appendix, we
(i) analyze the computational cost of dataset generation and the training process in the two cases of planner-based supervision and verifier-reward supervision, and
(ii) quantify their difference using asymptotic complexity.

\paragraph{Parameters and Notation.}
Let $n_i$ denote the size of the $i$-th planning instance and $N$ the number of problems in the dataset.
Let $L_i$ be the length of the corresponding plan, $E$ the number of training epochs, and $G$ the number of candidate plans sampled per problem during GRPO.
Let $T_i = T_{\text{in}}(n_i) + T_{\text{out}}(n_i)$ be the total number of tokens associated with the $i$-th problem encoding and its plan.
Furthermore, let us indicate as $P$ the number of trainable parameters of the language model.
We denote by $C_{\text{planner}}(n_i)$ the cost of solving the $i$-th instance using a classical planner, by $C_{\text{validation}}(n_i, L_i)$ the cost of validating a plan using a tool such as VAL, and by $C_{\text{generation}}(n_i)$ the cost of generating all domain-problem-plan tuples.
Finally, we denote by $C_{\text{language\_model}}(P, T_i)$ the computational cost of a forward pass of the language model and by $C_{\text{update}}(P, T_i)$ the cost of performing gradient
computation and parameter updates for a single training example.

\paragraph{Inference Cost.}
The cost of generating a single candidate plan (inference) for the $i$-th instance is:
\[
C_{\text{inference},i} = O\big(C_{\text{language\_model}}(P, T_i)\big).
\]
Therefore, the total inference cost over the entire dataset is:
\[
C_{\text{inference}}^{(N)}
=
O\!\left(\sum_{i=1}^{N} C_{\text{language\_model}}(P, T_i)\right).
\]

\paragraph{Planner-Based Training.}

In supervised or planner-based training, each dataset element requires both problem generation and a reference plan generated by a planner.
The total dataset generation cost is:
\[
\begin{aligned}
C_{\text{planner:data-generation}}
&=
O\!\left(\sum_{i=1}^{N} \big(C_{\text{generation}}(n_i) + C_{\text{planner}}(n_i)\big)\right) \\
&\approx
O\!\left(\sum_{i=1}^{N} C_{\text{planner}}(n_i)\right),
\end{aligned}
\]
because the planner cost dominates problem emission.

Supervised fine-tuning processes the entire dataset for $E$ epochs.
The total training cost includes both inference and parameter updates:
\[
C_{\text{planner:training}}
=
O\!\left(
E \cdot \sum_{i=1}^{N}
\big(C_{\text{language\_model}}(P, T_i) + C_{\text{update}}(P, T_i)\big)
\right).
\]
Combining dataset generation and training, the total cost is:
\[
C_{\text{planner:total}}
=
O\!\left(\sum_{i=1}^{N} C_{\text{planner}}(n_i)\right)
+
O\!\left(
E \cdot \sum_{i=1}^{N}
\big(C_{\text{language\_model}}(P, T_i) + C_{\text{update}}(P, T_i)\big)
\right).
\]

\paragraph{Verifier-Reward Training.}

In verifier-reward training, no reference plans are generated.
Therefore, the dataset generation cost reduces to:
\[
C_{\text{rl:data-generation}}
=
O\!\left(\sum_{i=1}^{N} C_{\text{generation}}(n_i)\right).
\]
During training, $G$ candidate plans are sampled per problem and evaluated using
a verifier.
Defining the total verification cost over the dataset as
\[
C_{\text{validation}}^{(N)} =
\sum_{i=1}^{N} O\left(C_{\text{validation}}(n_i, L_i)\right),
\]
the total training cost becomes:
\[
C_{\text{rl:training}}
=
O\!\left(
E \cdot G \cdot
\sum_{i=1}^{N}
\big(
C_{\text{language\_model}}(P, T_i)
+
C_{\text{validation}}(n_i, L_i)
+
C_{\text{update}}(P, T_i)
\big)
\right).
\]
Combining dataset generation and training, the total cost is:
\[
C_{\text{rl:total}}
=
O\!\left(\sum_{i=1}^{N} C_{\text{generation}}(n_i)\right)
+
O\!\left(
E \cdot G \cdot
\sum_{i=1}^{N}
\big(
C_{\text{language\_model}}(P, T_i)
+
C_{\text{validation}}(n_i, L_i)
+
C_{\text{update}}(P, T_i)
\big)
\right).
\]

\paragraph{Implications.}

The verifier-reward training process replaces an upfront cost proportional to $\sum_{i=1}^{N} C_{\text{planner}}(n_i)$ with costs proportional to $\sum_{i=1}^{N} C_{\text{generation}}(n_i)$ for dataset construction, and $E \cdot G \cdot C_{\text{validation}}^{(N)}$ during the training process.
Since plan validation is \textit{polynomial} in plan length, and substantially cheaper than full planning, this shift enables scaling to larger datasets without requiring planner-generated supervision.
This theoretical analysis is corroborated by our empirical findings.
On the one hand, training based on verifier-reward incurs a substantially higher computational cost per epoch due to repeated sampling, validation, and gradient updates, which results in longer wall-clock training times.
On the other hand, it accelerates convergence in terms of training epochs, therefore reaching performance saturation in significantly fewer epochs than supervised
fine-tuning.

These observations suggest several directions for future work.
Herewith, we highlight two of them, on the technical side of the spectrum.
\textit{First}, the choice of the supervised bootstrapping checkpoint could be optimized to maximize  efficiency gains of Reinforcement Learning.
\textit{Second}, because verifier-reward training provides a more informative learning signal \textit{per example}, in principle it may be possible to reduce the size of the training dataset while maintaining comparable performance, therefore further reducing the per-step
training cost.